\documentclass[10pt,twocolumn,letterpaper]{article}

\usepackage{iccv}
\usepackage{times}
\usepackage{epsfig}
\usepackage{graphicx}
\usepackage{amsmath}
\usepackage{amssymb}

\usepackage[font={small},skip=1pt]{caption} %
\usepackage{pifont}%
\newcommand{\cmark}{\ding{51}}%
\newcommand{\xmark}{\ding{55}}%
\usepackage{array}
\usepackage{ragged2e}
\usepackage{textcomp} %
\usepackage{gensymb}
\usepackage[inline]{enumitem}
\usepackage{bm}
\usepackage[normalem]{ulem}
\usepackage{siunitx}
\usepackage{subcaption}
\usepackage{tabularx}
\usepackage{tabu}
\graphicspath{ {images/} }
\usepackage[numbers,sort]{natbib} %

\newcommand\norm[1]{\left\lVert#1\right\rVert}

\newcolumntype{L}[1]{>{\raggedright\let\newline\\\arraybackslash\hspace{0pt}}m{#1}}
\newcolumntype{R}[1]{>{\raggedleft\let\newline\\\arraybackslash\hspace{0pt}}m{#1}}
\newcolumntype{C}[1]{>{\centering\let\newline\\\arraybackslash\hspace{0pt}}m{#1}}

\def\vec#1{\mathchoice{\mbox{\boldmath $\displaystyle\bf#1$}}
{\mbox{\boldmath $\textstyle\bf#1$}}
{\mbox{\boldmath $\scriptstyle\bf#1$}}
{\mbox{\boldmath $\scriptscriptstyle\bf#1$}}}
\def\vaz #1{\protect\vec #1}

\def\vecs#1{\mathchoice%
	{\mbox{\boldmath \small $\displaystyle\bf#1$}}
	{\mbox{\boldmath \small $\textstyle\bf#1$}}
	{\mbox{\boldmath \small $\scriptstyle\bf#1$}}
	{\mbox{\boldmath \small $\scriptscriptstyle\bf#1$}}}
\def\vsaz #1{\protect\vecs #1}

\def\mat#1{\mathchoice{\mbox{\boldmath$\displaystyle\tt#1$}}
{\mbox{\boldmath$\textstyle\tt#1$}}
{\mbox{\boldmath$\scriptstyle\tt#1$}}
{\mbox{\boldmath$\scriptscriptstyle\tt#1$}}}
\def\maz #1{\protect\mat #1}

\usepackage[linktocpage,breaklinks=true,bookmarks=false]{hyperref}
\usepackage{xcolor}
\hypersetup{
    colorlinks,
    linkcolor=blue,
    citecolor=red,
    urlcolor=blue
}

\iccvfinalcopy %

\def\iccvPaperID{22} %

\ificcvfinal\fi

\begin{document}

\title{A Geometric Approach to Obtain a Bird's Eye View From an Image}

\author{Ammar Abbas~\thanks{The author is now at Latent Logic, Oxford}\\
VGG, Dept.\ of Engineering Science\\
University of Oxford\\
United Kingdom\\
{\tt\small sabbas.ammar@gmail.com}
\and
Andrew Zisserman\\
VGG, Dept.\ of Engineering Science\\
University of Oxford\\
United Kingdom\\
{\tt\small az@robots.ox.ac.uk}
}

\maketitle
\ificcvfinal\thispagestyle{empty}\fi

\begin{abstract}
   {
The objective of this paper is to rectify any monocular image by
computing a homography matrix that transforms it to a geometrically correct bird's eye (overhead) view. 

We make the following contributions: (i) we show that the homography matrix can
be parameterised with only four geometric parameters that specify the horizon line and the vertical vanishing point,  or only
 two if the field of view or focal length is known; (ii)  We introduce
a novel representation for the geometry of a line or 
point (which can be at infinity) that is suitable for regression with a convolutional neural network (CNN);
(iii) We
introduce a  large synthetic image dataset with ground truth for the
orthogonal vanishing points, that can be used  for training a
CNN to predict these geometric
entities; and finally (iv) We achieve state-of-the-art results on horizon detection,
with 74.52\% AUC on the \emph{Horizon Lines in the Wild} dataset.
Our method is fast and robust, and can be used
to remove perspective distortion from videos in real time. The code is available at: {\small \url{https://github.com/SAmmarAbbas/birds-eye-view} } }
\end{abstract}

\section{Introduction}

\begin{figure}
\centering
\includegraphics[width=0.9\columnwidth]{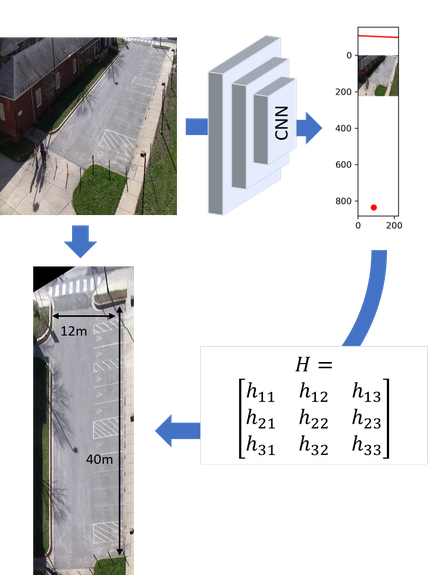}
\caption{\small An overview of our method for obtaining the bird's eye view of a scene from a single perspective image. A CNN is used to estimate the vertical vanishing point and ground plane vanishing line (horizon) in the image, as shown by the red dot and line in the example. This point and line in turn determine the homography matrix, $\maz H$,  that maps the image to the overhead view with perspective distortion removed. Measurements on the ground plane (up to an overall scale) can then be made directly on the rectified image.}
\label{fig:main_fig}
\end{figure}

Understanding the 3D layout of a scene from a single perspective image is one 
of  the fundamental problems in computer
vision. Generating a bird's eye (or overhead, or orthographic) view of the scene plays a part in this understanding
as it allows the perspective distortion of the ground plane to be removed. This {\em rectification} of the ground
plane allows the scene geometry on the ground plane to be measured directly from an image. 
It can be used as a
pre-processing step for many other computer vision tasks like object
detection~\citep{redmon2017yolo9000,he2017mask} and 
tracking~\citep{danelljan2015convolutional}, and has 
applications in video surveillance and traffic control. For example, in 
crowd counting, where perspective distortion
affects the crowd density in the image, the crowd density can instead be predicted in the 
world~\citep{liu2018geometric}.

Since obtaining a bird's eye view from an image involves computing a
rectifying planar homography, it might be thought that the most direct
way to obtain this transformation would be to regress the eight
parameters that specify the homography matrix. Instead, we show that
this homography can be parametrised with only four parameters
corresponding to the vertical vanishing point and ground plane vanishing line (horizon) in the image,
and that these geometric entities can be regressed directly using a
Convolutional Neural Network (CNN). Furthermore if the focal length of
the camera is known (or equivalently the field of view) from the EXIF
data of the image, then only two further parameters are required
(corresponding to the vanishing line). We show that given these
minimal parameters, the homography matrix
that transforms the source image into the
desired bird's eye view
can be composed through a sequence
of simple transformations. Furthermore, the geometric entities can also be used directly for scene 
understanding~\citep{fouhey2015single}.

For the purpose of training a CNN, we
introduce and release~\footnote{\url{https://drive.google.com/open?id=1o9ydKCnh0oyIMFAw7oNxQohFa0XM4V-g}} the largest up-to date dataset which contains the ground
truth values for all the three orthogonal vanishing points with the
corresponding internal camera matrices, and tilt and roll of the camera for
each image. We also propose a novel representation for the
geometry of vanishing lines and points in image space, which handles the standard challenge that these
entities can lie within the image but also can be very distant from the image, making this representation a good choice for the network prediction.

In summary, we make the following four contributions: (i) we propose a
minimal parametrisation for the homography that maps to the bird's eye
view. This requires only four parameters to be specified (the
vanishing point and vanishing line), or only two if the focal length
of the camera is known (the vanishing line); (ii) we
propose a new geometric representation for encoding vanishing points and lines that is suitable for neural network computation; (iii) we generate and
release a large synthetic dataset, CARLA-VP, that can be used for
training a CNN model to predict vanishing points and lines from a
single image; and (iv) we show that a CNN trained using our four scalar
parameterisation exceeds the performance of the state of the art on
standard real image benchmarks for
horizon detection~\citep{workman2016horizon}.

We also show that current methods~\citep{lezama2014finding,he2016deep} can fail for horizon prediction when
the actual horizon line lies outside of the image. 
This failure is due to the parameterization
used,  as well as to the training data used (which mostly contains horizon lines
inside the image since it is easier to annotate them). We avoid this annotation problem
by using synthetic data for training, where images can be
generated following any desired distribution and the annotations are
more precise as well. We compare to results on a benchmark dataset~\citep{virat2011dataset} 
in section~\ref{subsec:comparison_third_party}.

\section{Related Work}
\label{sec:related_work}

\paragraph{Estimating homographies:}

\citet{bruls2018right} use GANs to estimate the bird's eye view;
however, since they don't enforce a pixel-wise loss, the geometry of
the scene may not be correctly recovered as they mention in failure
cases. Moreover, they train and test only on first person car driver
views~\citep{maddern20171} where some assumptions can be made
(pitch$\approx$0, roll$\approx$0). \citet{liu2018geometry} pass an
additional relative pose to a CNN for view synthesis which contains
information about the relative 3D rotation, translation and camera
internal parameters. 

\vspace{-10pt}
\paragraph{Estimating the focal length of the camera:}
\label{subsec:methods_focal_length}

One of the ways to calculate focal length is by estimating the field
of view from the image. The focal length $f$ is inversely related to
the field of view $\gamma$ of the camera given constant image width
$w$ as:
\begin{equation}
\label{eq:focal_length_fov}
\tan(\frac{\gamma}{2}) = \frac{\frac{w}{2}}{f}
\end{equation}
\citet{workman2015deepfocal} use this approach
to predict a camera's focal length by estimating the field of view
directly from an image using a CNN. However, since they only predict
horizontal field of view, they assume that the camera has equal focal
length on both the axes which may not be true. In addition, based on
the findings in ~\citep{he2018learning}, we know that predicting the field
of view directly from an image can be a challenging task since two
similar looking images may have large differences in field of view. We
estimate the focal length of the camera from the horizon
line and the vertical vanishing point (and describe the advantages 
in section~\ref{subsec:ablation_study}).

\vspace{-10pt}
\paragraph{Computing vanishing points and lines:}

One simple way to estimate the horizon line or the vertical vanishing
point is by finding the intersection point of the lines in the image
which belong to the orthogonal directions in the image. More
specifically, this could involve using a Hough
transform on the detected lines to vote among
the candidate vanishing points~\cite{Tuytelaars98}, and many other voting schemes have
been investigated~\cite{collins1990vanishing}, including weighted voting~\cite{shufelt1999performance}
and expectation maximization~\cite{antone2000automatic}.
More recently, ~\citet{lezama2014finding} 
vote both in the image domain
and PClines dual
spaces~\citep{dubska2011pclines}.
The above methods have a limitation as they rely on line detection as
the core step and may fail when the image does not have lines in the
major directions. Fortunately, there are other important cues in an
image which help us to estimate the horizon line or the vanishing
points such as colour shifts, distortion in objects' shapes, change
in texture density or size of objects around the image \etc and that is where deep learning methods can help us.

\vspace{-10pt}
\paragraph{Datasets:} 
There are a few existing datasets which contain the ground truth for
the three orthogonal vanishing points in the scene namely,
\emph{Eurasian Cities} dataset~\citep{barinova2010geometric},
\emph{York Urban} dataset~\citep{denis2008efficient} and the
\emph{Toulouse Vanishing Points}
dataset~\citep{angladon2015toulouse}. However, these datasets contain
only around 100 images in total.
\citet{borji2016vanishing} propose a CNN based method
which is trained  by annotating
vanishing points in YouTube frames. 
Recently, \citet{workman2016horizon} collected a new dataset called
\emph{Horizon Lines in the Wild} (HLW) which contains around 100K images
with ground truth for the horizon line. 
However, their dataset mostly contains images where  the horizon line lies within the image, and
does not
contain explicit labeling for the orthogonal vanishing points.
Because of the unavailability of a large dataset which contains the
orthogonal vanishing points, we generate a
large-scale synthetic dataset that contains the required ground
truth. This allows us to train a CNN to predict these  geometric
entities. We discuss this in detail in section~\ref{sec:dataset}

\section{Predicting a homography from the horizon line and the vertical vanishing point}
\label{sec:predict_homography_from_horvpz}

In the following we assume that we know the 
vertical vanishing point and horizon line in the
image,  and show geometrically how these are used to compute the rectifying homography matrix.
In section~\ref{sec:predict_horvpz} we describe how to
estimate these geometric entities using a CNN.

\begin{figure}[!tb]
\centering
\begin{subfigure}{.9\columnwidth}
  \centering
  \includegraphics[width=\columnwidth]{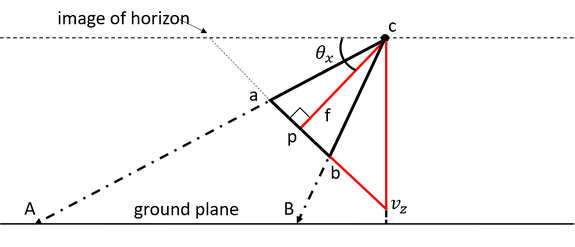}
  \caption{\label{subfig:before_rotate}}
\end{subfigure}
\begin{subfigure}{.9\columnwidth}
  \centering
  \includegraphics[width=\columnwidth]{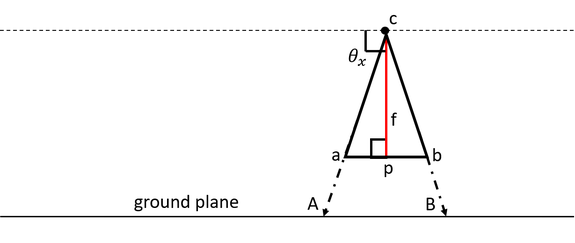}
  \caption{\label{subfig:after_rotate}}
\end{subfigure}
\vspace*{10pt}\caption{ Side view of a camera viewing a scene. (\subref{subfig:before_rotate}) The camera is tilted with an angle $\theta_x$. Its centre is represented by $c$ and $f$ is the focal length. $\overline{ab}$ is the image plane, $p$ the principal point, and $v_z$  the vertical vanishing point. (\subref{subfig:after_rotate}) The camera is rotated to look directly vertically down on the scene -- the 
bird's eye view.}
\label{fig:camera_side_view}
\end{figure}

The method involves applying a sequence of projective transformations
to the image that are equivalent to rotating the camera and
translating the image in order to obtain the desired bird's eye
view. As shown in figure~\ref{fig:camera_side_view} the key step is to
use the horizon line to determine the rotation required, but in order
to know the rotation angle from the horizon we require the camera
internal calibration matrix. Assuming that the camera has square
pixels (zero skew) and that the principal point is at the centre
of the image, then the only unknown parameter of the internal calibration matrix
is the focal length, and this can be determined once both the vertical
vanishing point and horizon are known as described below.

\begin{figure*}[tb]
\centering
\begin{subfigure}{.19\textwidth}
  \centering
  \includegraphics[width=\columnwidth]{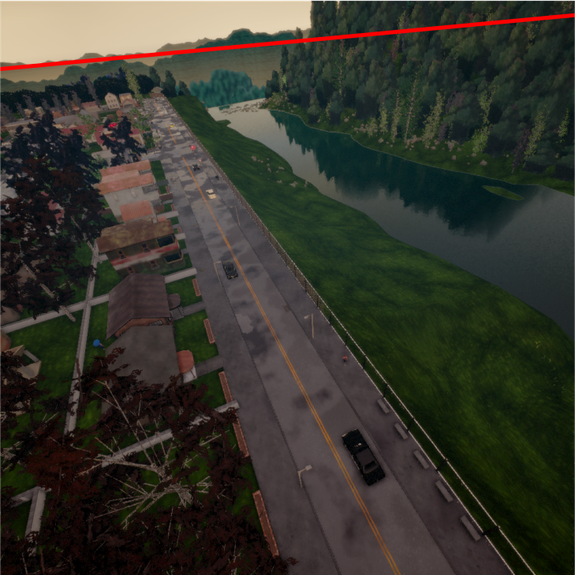}
    \caption*{Original Image}
\end{subfigure}%
\hspace{1pt}
\begin{subfigure}{.19\textwidth}
  \centering
  \includegraphics[width=\columnwidth]{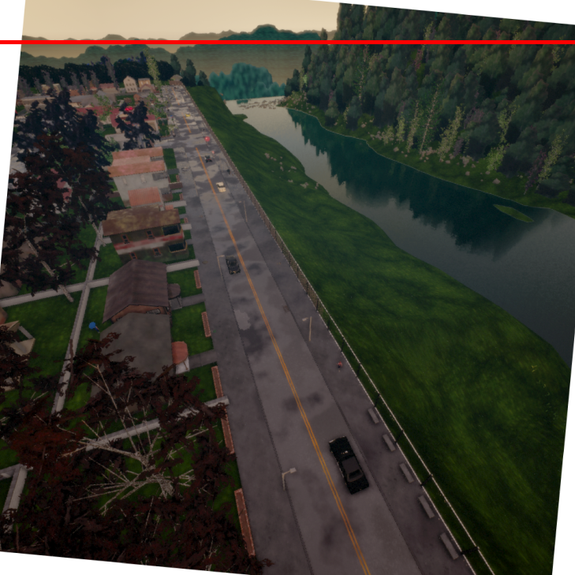}
  \caption*{Step A}
\end{subfigure}
\begin{subfigure}{.19\textwidth}
  \centering
  \includegraphics[width=\columnwidth]{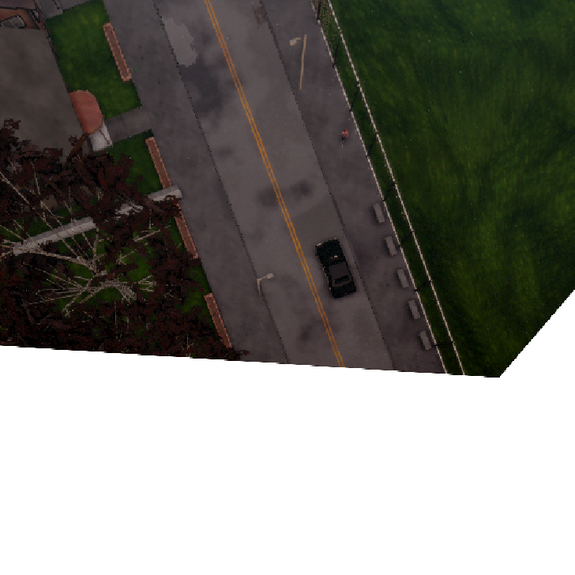}
  \caption*{Step B}
\end{subfigure}
\begin{subfigure}{.19\textwidth}
  \centering
  \includegraphics[width=\columnwidth]{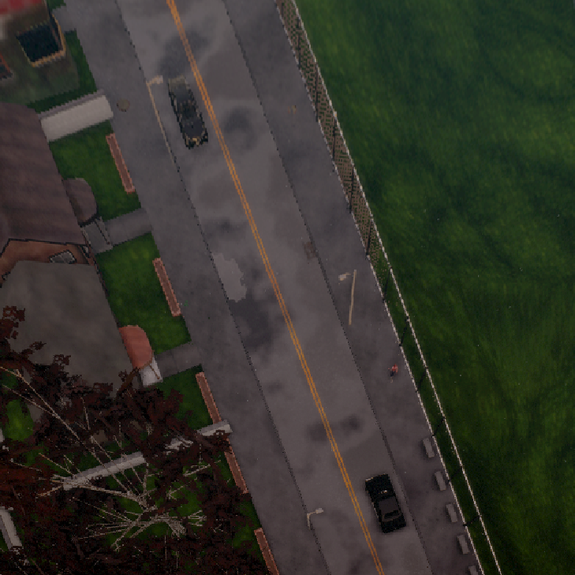}
  \caption*{Step C}
\end{subfigure}
\begin{subfigure}{.19\textwidth}
  \centering
  \includegraphics[width=\columnwidth]{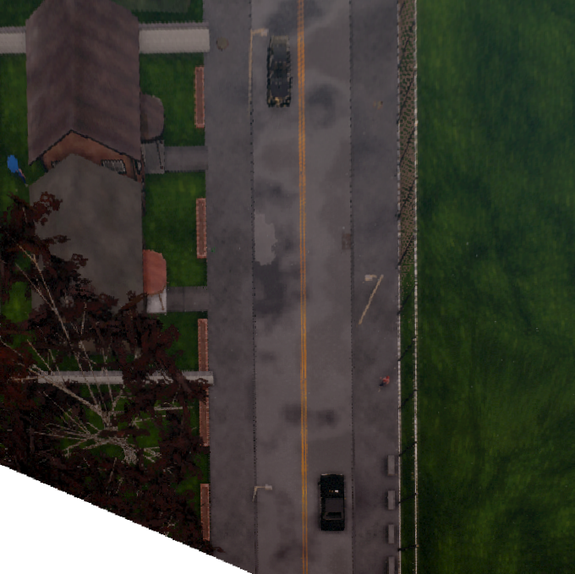}
  \caption*{Step D}
\end{subfigure}
\vspace*{10pt}\caption{\small Step-by-step transformations of the first image to obtain the desired bird's eye view. The different sub-figures correspond to images obtained after performing the steps described in section~\ref{sec:predict_homography_from_horvpz} for predicting the homography matrix.}
\label{fig:steps_transformation}
\end{figure*}

\vspace{-5pt}
\paragraph{Preliminaries.} We will use the following relationship~\citep{hartley2003multiple} between 
image coordinates before and after a rotation of the camera about its centre:
\vspace{-5pt}
\begin{equation}
\vaz x' = \maz K \maz R \maz K^{-1}  \vaz x   
\label{eq:rotation_camera}
\end{equation}
where $\vaz x$
represents image pixels for scene coordinates $\vsaz X$ before the camera
rotation, and  $\vaz x'$ are the resultant image pixels for the same scene coordinates $\vsaz X$ after the rotation, 
and the internal calibration matrix $\maz K$ is given by

\vspace{-5pt}
\begin{equation}
    \maz K = 
    \begin{bmatrix}
    	f & 0 & w/2 \\
    	0 & f & h/2 \\
    	0 & 0 & 1
    \end{bmatrix}
\end{equation}

where $f$ is the focal length of the camera, $w$ is the width of the image, and $h$ the height of the image.

To compute the matrix $\maz K$, we only need to
find the focal length $f$ of the camera. 
As explained in~\citep{hartley2003multiple} the focal length can be obtained directly from the 
relationship
\vspace{-5pt}
\begin{equation}
\vaz h = \maz \omega \vaz v
\label{eq:pole_polar_horvpz}
\end{equation}
where $\vaz h$ is the horizon line  and $\vaz v$ the vertical
vanishing point, and $\maz \omega$ is known as the \emph{image of absolute conic} 
which is unaffected by the camera rotation and is given by $\maz \omega = (\maz K \maz K^T)^{-1}$.

The rotation matrix $\maz R$ in equation~\eqref{eq:rotation_camera} can be
composed of rotations about different axes. We use this property to
first rotate the camera about  its principal axes to correct for the roll
of the camera, and then about the x-axis of the camera to reach an overhead view of the scene.
We next describe the sequence of projective transformations.

\vspace{-10pt}
\paragraph{Step A: removing camera roll.} 
The first step is to apply a rotation about the principal axis to remove any camera roll,
 so that the camera's x-axis is
parallel to the X-axis of the world. See step A in figure~\ref{fig:steps_transformation} for its effect. The
roll of the camera is computed from the horizon line. Given a horizon line of the
form $ax+by+c=0$, the roll of the camera $\theta_z$ is given by
$    \theta_z = \tan^{-1}(\frac{-a}{b})$.
The rotation matrix $R_{roll}$ for rotating about the principal axis is computed using
$\theta_z$. 

\vspace{-10pt}
\paragraph{Step B: removing camera tilt.} 
The next step is to rotate about the camera x-axis to remove the
camera tilt.  See step B in figure~\ref{fig:steps_transformation}
for its effect.  The rotation matrix for rotation about the camera x-axis 
requires only one parameter which is the camera tilt
$\theta_x$. The camera tilt can be found from the focal length and one
of the geometrical entities, either the horizon line or the vertical vanishing
point. Given the focal length of the camera $f$ and the perpendicular
distance from the vertical vanishing point to the principal point
$\norm{v_z}$, we can find tilt $\theta_x$ of the camera as 
$    \theta_x = \frac{\pi}{2}-\tan^{-1}(\frac{\norm{v_z}}{f})$.
See figure~\ref{fig:camera_side_view} for the corresponding notation. 
At this point, the homography matrix $\maz H_{rot}$ is given as:
\begin{equation}
   \maz  H_{rot} = \maz K \maz R_{tilt} \maz K^{-1} \maz R_{roll}
\label{eq:only_rot_homography}
\end{equation}
where $\maz R_{tilt}$ is the rotation matrix for rotating about the x-axis to recover the camera tilt.

\vspace{-10pt}
\paragraph{Step C: image translation.} 
Once we have the effect of camera rotation, we also need to translate
the camera so that it is directly above the scene and captures the
desired bird's eye view. We achieve this by applying $H_{rot}$ to the
four corners of the source image which returns the corresponding
corners for the destination image. We define a translation matrix
which maps the returned corners to the corners of our final canvas,
thereby giving us the full view of the scene from above. See step C in figure~\ref{fig:steps_transformation}.

\vspace{-10pt}
\paragraph{Step D: optional rotation.} 
We also have an optional step which can be seen in step D in
figure~\ref{fig:steps_transformation}. It deals with aligning the
major directions in the image with the axes in the Cartesian
coordinate system by rotating the final image by an angle
$\theta_{align}$. This angle can be obtained from one of the principal
horizontal vanishing points that relates to one of the major
directions in the image. We show in
section~\ref{sec:represent_geometry} how to represent this vanishing
point by a single scalar.

\noindent In summary, the steps of the algorithm are:
\begin{itemize}
\itemsep0em
    \item Calculate the focal length of the camera using the predicted horizon line and the vertical vanishing point from a single image.
    \item Calculate the camera roll from the horizon line which gives us $\maz R_{roll}$.
    \item Calculate the camera tilt from the focal length and the vertical vanishing point which in turn is used to calculate $R_{tilt}$
    \item Calculate the translation matrix $T_{scene}$ using the homography matrix $H_{rot}$ from eq.~\ref{eq:only_rot_homography} to map the corners of the image to the destination image.
    \item \emph{(Optional)} Calculate $R_{align}$ from the principal horizontal vanishing point in the scene.
    \item Compose all above transformation matrices together to calculate the final homography matrix which is given as follows:
    
\end{itemize}

\vspace{-10pt}
\begin{equation}
   \maz  H = \maz R_{align} \maz T_{scene} \maz K \maz R_{tilt} \maz K^{-1} \maz R_{roll} 
\label{eq:overall_homography}
\end{equation}

\section{Predicting the horizon line and the vertical vanishing point}
\label{sec:predict_horvpz}

In this section we describe how the geometric entities are represented in a form suitable for regression with
a CNN.  The key point is that the entities can be at infinity in the image plane (e.g.\ if the camera is facing down then
the vanishing line is at infinity) and so a representation is needed to map these entities to finite points for the CNN prediction. To achieve
this we borrow ideas from the standard stereographic projection used to obtain a map of the earth~\cite{snyder1997flattening}.

\begin{figure}
\centering
\begin{subfigure}{.6\columnwidth}
\centering
\includegraphics[width=\columnwidth]{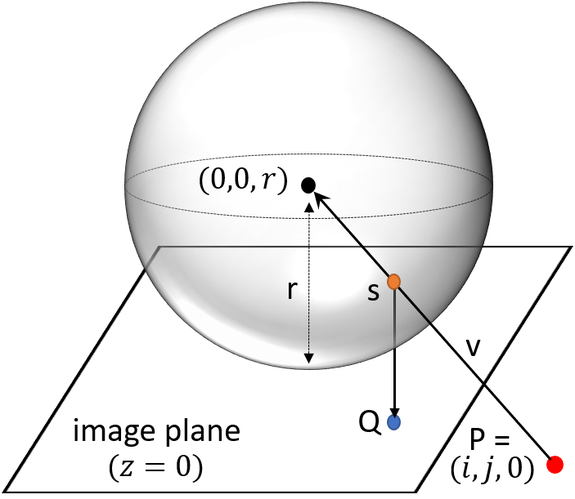}
\caption{\label{subfig:point_proj_2sphere}}
\end{subfigure}

\vspace*{10pt}\begin{subfigure}{\columnwidth}
\centering
\includegraphics[width=\columnwidth]{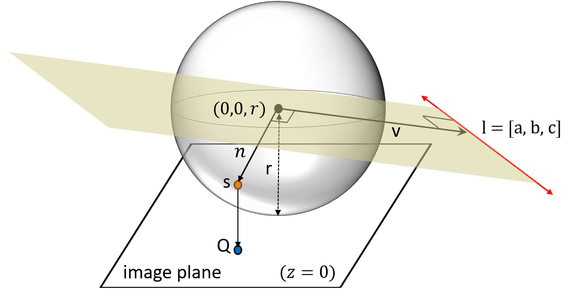}
\caption{\label{subfig:line_proj_2sphere}}
\end{subfigure}
\vspace*{10pt}\caption{Representation of the geometry of a vanishing point and a horizon line. (\subref{subfig:point_proj_2sphere}) A point $P$ (which could be a point at infinity) is represented by a finite point $Q$ on the 2D plane. As the point 
$Q$ is projected from the intersection point $s$ on the lower hemisphere, it is constrained to lie within a circle of radius $r$ on the image plane. (\subref{subfig:line_proj_2sphere}) A line $l$ is represented by a finite point $Q$ on the image plane. The plane connects the centre of the sphere with the line $l$. Point $s$ is the intersection of the plane normal with the sphere boundary,  and $Q$ is the projection of $s$ on the image plane. }
\label{fig:proj_2sphere}
\end{figure}

\subsection{Representing the geometry of the horizon line and the vanishing points}
\label{sec:represent_geometry}

We first describe the representation method for a point. See
figure~\ref{fig:proj_2sphere} for the notation introduced
ahead. Suppose there is a sphere of
radius $r$ which is located at point $(0,0,r)$, and
let the image plane be at $z=0$. 
Then we can draw a
line connecting any point $P$ on the image plane  to the sphere
centre. The point $s$ on sphere where this line intersects
the sphere is given by $ s = r \frac{\bm{v}}{\norm{\bm{v}}} $ where $v$ is a vector from the sphere centre to $P$ and
$s$ is a 3-D point on sphere.  Finally, we project the point
$s$ onto the image plane at $Q$ 
using orthogonal projection. This effectively allows us to represent any 2D point $P$ on the image
by a point $Q$ in a finite domain (within a circle of radius r), irrespective of whether the original point $P$ is finite or at infinity.

For a line $l$, we take a slightly different approach to represent its
geometry. We draw a plane which connects the line $l$ to the centre of
the sphere. There is a one-to-one mapping between the line $l$ and the
plane drawn corresponding to it. The normal $n$ to the
plane from the sphere centre intersects the surface of the sphere
in the lower hemisphere at a point $s$. Once again, we orthogonally project this point $s$ onto the
plane. This gives a unique finite point representation for any line $l$ in the
infinite plane. In this way, we can represent the horizon line and the
vertical vanishing point in the image by a total of four scalars which
lie in the range $[ -r,r ]$.

The optional principal horizontal vanishing point can be
represented by a single scalar. We know that the horizontal vanishing
points lie on the horizon line, so we only need to measure its
position on the horizon line. We do so by measuring the angle between two
vectors: a vector $\vec{v_1}$ which goes from the sphere centre $C$ to the
required horizontal vanishing point and another vector $\vec{v_2}$
which is normal to the horizon from $C$.

\section{The CARLA-VP Dataset}
\label{sec:dataset}

There is no large scale dataset with ground truth for
the horizon line and the vertical vanishing point
available for training a CNN, so here we generate a synthetic training dataset.
Table~\ref{table:comp_datasets} gives statistics on relevant datasets.

\begin{table}[tb]
\small
\begin{center}
\begin{tabular}{|L{75pt}|R{32pt}|R{39pt}|R{39pt}|}
\hline
\multicolumn{1}{|m{75pt}|}{\centering Dataset} & \multicolumn{1}{m{32pt}|}{\centering Training} & \multicolumn{1}{m{39pt}|}{\centering Validation} & \multicolumn{1}{m{39pt}|}{\centering Test} \\
 \hline
 HLW~\citep{workman2016horizon} & 100553 & 525 & 2018 \\ 
 VIRAT Video~\citep{virat2011dataset} & - & - & 11 videos \\
CARLA-VP  & 12000 & 1000 & 1000 \\
 \hline
\end{tabular}
\end{center}
\caption{Comparison of the number of examples  in training, validation, and test set for different datasets. Note: The videos for VIRAT dataset aren't divided into different training, validation or test sets by the publishers.}
\label{table:comp_datasets}
\end{table}

\subsection{Synthetic dataset}

We use CARLA~\citep{Dosovitskiy17} which is an open-source simulator
built over the Unreal Engine~4
to create our dataset. It generates photo-realistic images with
varying focal length, roll, tilt and height of the camera in various
environments.

We choose a uniformly random value for the height of the camera
ranging from a ground person's height to around 20 metres. We also
choose a uniformly random value for tilt of the camera in the range
$(\ang{0}, \ang{40}]$. We choose a value for camera roll from a normal
distribution with $\mu=\ang{0}$ and $\sigma=\ang{5}$ which is
truncated in the range $[\ang{-30}, \ang{30}]$.

\begin{figure}[!tb]
\centering
\begin{subfigure}{.24\columnwidth}
  \centering
  \includegraphics[width=\columnwidth,height=\columnwidth]{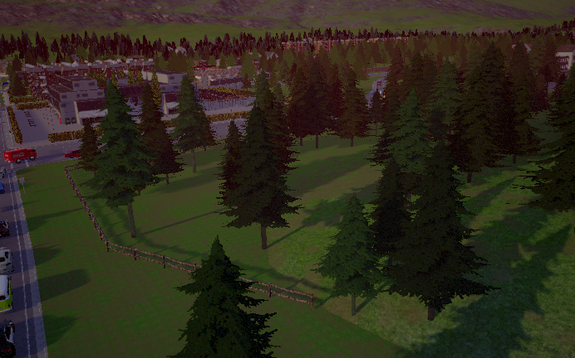}
\end{subfigure}
\begin{subfigure}{.24\columnwidth}
  \centering
  \includegraphics[width=\columnwidth,height=\columnwidth]{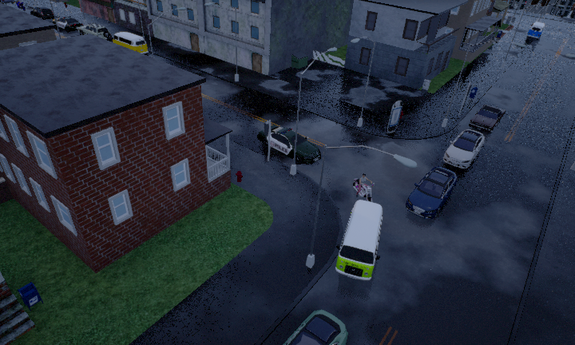}
\end{subfigure}
\begin{subfigure}{.24\columnwidth}
  \centering
  \includegraphics[width=\columnwidth,height=\columnwidth]{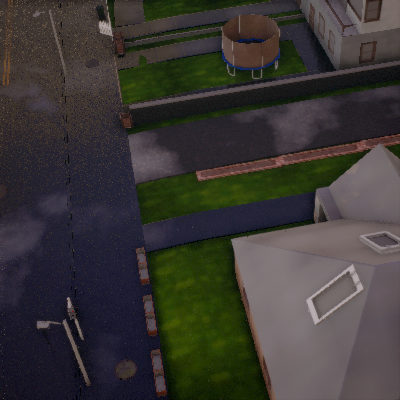}
\end{subfigure}
\begin{subfigure}{.24\columnwidth}
  \centering
  \includegraphics[width=\columnwidth,height=\columnwidth]{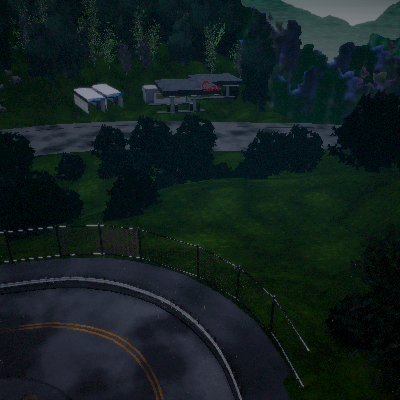}
\end{subfigure}

\centering
\begin{subfigure}{.24\columnwidth}
  \centering
  \includegraphics[width=\columnwidth,height=\columnwidth]{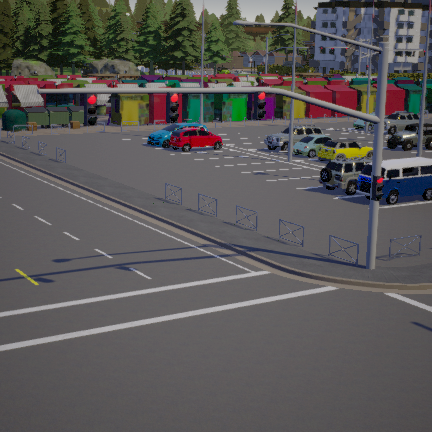}
\end{subfigure}
\begin{subfigure}{.24\columnwidth}
  \centering
  \includegraphics[width=\columnwidth,height=\columnwidth]{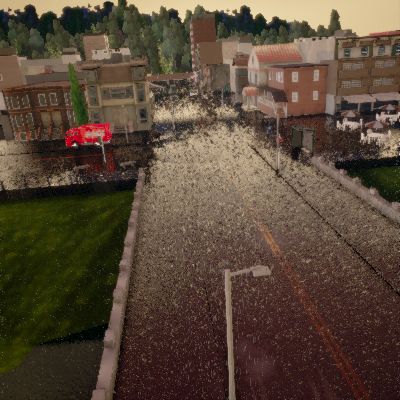}
\end{subfigure}
\begin{subfigure}{.24\columnwidth}
  \centering
  \includegraphics[width=\columnwidth,height=\columnwidth]{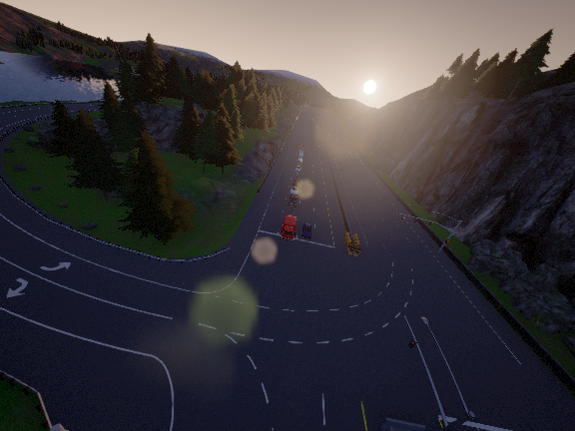}
\end{subfigure}
\begin{subfigure}{.24\columnwidth}
  \centering
  \includegraphics[width=\columnwidth,height=\columnwidth]{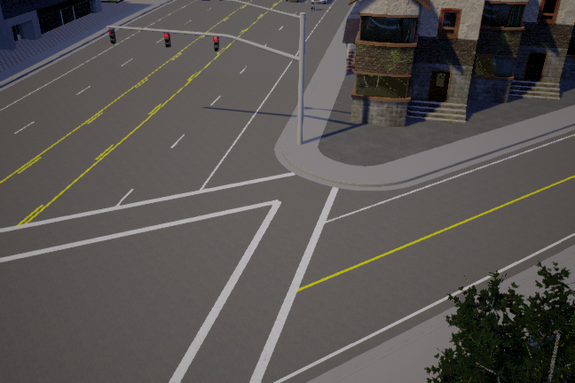}
\end{subfigure}
\caption[Sample images from the CARLA-VP dataset]{Sample images from the CARLA-VP dataset. The images show a wide variety of settings i.e.\ different camera positions and orientations, different weathers and different times of the day.}
\label{fig:samples_dataset}
\end{figure}

CARLA provides the ability to change the field of view of the
camera. This allows us to effectively change the focal length of the
camera as given in equation~\eqref{eq:focal_length_fov}. We use a
uniformly random value for field of view from the range $[\ang{15},
\ang{115}]$ which is carefully selected based on the images that are
generally captured or are obtained from traffic surveillance
cameras. The different parameters that we have discussed above allow
us to generate a wide variety of images with different aspect ratios that resemble real-world
images. We will refer to this dataset as \emph{CARLA-VP} (i.e.\ CARLA with Vanishing Points).
See figure~\ref{fig:samples_dataset} for a few samples from the dataset.

\subsection{Ground Truth Generation}

Synthetic datasets allow us to create precise ground truths. We
mentioned above that we can change tilt, roll or yaw of the camera in
the CARLA simulator. This gives us the value for the camera's rotation
matrix $\maz R = \left[ \begin{array}{ccc} \vaz r_1 & \vaz r_2 & \vaz r_3
 \end{array} \right]$ by composing it as a composition of
individual rotation matrices. Similarly, we also know the internal calibration
matrix $\maz K$ of the camera as CARLA uses a simplified form and we
already know the focal length~\eqref{eq:focal_length_fov}.

Using $\maz K$ and $\maz R$, we can generate ground truth for the orthogonal
vanishing points. Consider a point at infinity in the z direction,  $\vaz z_{\infty}$,
which is represented as $\begin{bmatrix} 0 & 0 & 1 & 0 \end{bmatrix}^T$
in homogeneous coordinates,  and its image  $v_z$. Then by the camera's
projection equation, we have:

\vspace{-10pt}
\begin{equation}
\vaz v_z =
\maz K
\left[
\begin{array}{c|c}
\maz R & \vaz t \\
\end{array}
\right]
\vaz z_{\infty}
=
\maz K
\left[
\begin{array}{ccc|c}
\vaz r_1 & \vaz r_2 & \vaz r_3 & \vaz t \\
\end{array}
\right]
\begin{bmatrix}
0 \\
0 \\
1 \\
0 
\end{bmatrix}
\nonumber
\end{equation}

\vspace{-5pt}
\begin{equation}
\label{eq:gt_vanishing_point}
\vaz v_z =
\maz K \vaz r_3
\end{equation}

Similarly, we can also solve for the orthogonal horizontal vanishing points in the scene which are given by 
$\vaz v_x = \maz K \vaz r_1$ and $\vaz v_y = \maz K \vaz r_2$, and the horizon line is given by
$\vaz h = \vaz v_x  \times \vaz v_y$. 

\section{Experiments}

In this section, we perform a range of experiments to evaluate our
method both qualitatively as well as quantitatively. We first explain
the performance measures and conduct an ablation study of the 
method in section~\ref{subsec:ablation_study}, where we also compare different 
CNN architectures.
We then evaluate our method on videos and compare its
performance quantitatively on the VIRAT Video dataset with some
qualitative results on the real-world images. Finally, we compare our
horizon detection method with previous state-of-the-art methods.

\subsection{Performance Measures}
\label{sec:performance_measure}

We use two performance measures. The first is the area
under the curve (AUC) metric
proposed by~\citet{barinova2010geometric} for evaluating
horizon line estimation.
For each test image sample,
the maximum
difference between the height of the ground truth and estimated
horizon over the image, divided by the image height, is computed;
and these values are then plotted for the test set,  where the x-axis represents the error percentage and
the y-axis represents the percentage of images having error less than
the threshold on the x-axis. The AUC is measured on this graph.

The second performance measure evaluates the camera internal and external parameters, in
particular the field of view (which depends on the predicted focal length), and the roll and tilt of the
camera. We measure the error in these parameters in degrees. Note, these quantities are not directly
estimated by the CNN, but are computed from the predicted vertical vanishing point and horizon line.

\subsection{Implementation details}

The final layer of the network is required to predict four scalars, and this is implemented using
regression-by-classification as
a multi-way softmax for each scalar over $b$ discretization bins. The
number of discretization bins is chosen as $b=500$ in our experiments.
An alternative would be to directly regress each scalar using methods similar 
to~\citep{fischer2015image,mahendran20173d}, but we did not pursue that here.

At test time, we consider the $c$  bins with the highest
probability, and use a weighted average of
these bins by their probabilities to calculate the regressed value. We find
that $c=11$ gives the best performance on the validation set.

The CNN is trained using TensorFlow~\citep{abadi2016tensorflow} v-1.8
in Python 3.6. It is initialized with  pre-trained weights from ImageNet
classification~\citep{deng2009imagenet}. All layers are
fine-tuned as the task at hand is inherently
different from the image classification task. We use the Adam
optimizer~\citep{kingma2014adam} 
with default parameters. The training starts with an initial
learning rate of 1e-2 which is divided by 10 up-to 1e-4 whenever the
validation loss increases. 

\subsection{Ablation Study}
\label{subsec:ablation_study}
        
\paragraph{Field of view vs vertical vanishing point.} We discussed in
section~\ref{sec:predict_horvpz} that our method for
calculating the bird's eye view involves estimating the internal and
external parameters of the camera. We do this by estimating the horizon
line and the vertical vanishing point from a given image. This
involves predicting four different scalars. However, we can further
reduce the number of parameters by predicting the field of view
instead of the vertical vanishing point. This is an even more compact
representation which uses only three scalars. It allows us to
calculate the focal length directly from the field of
view as in~(\ref{eq:focal_length_fov})~\citep{he2018learning}, and the tilt and roll of the camera from
the horizon line and focal length.

\begin{table}
\small
\begin{center}
\begin{tabular}{|L{85pt}|R{33pt}|R{33pt}|R{33pt}|}
\hline
\multicolumn{1}{|m{85pt}|}{\centering Model Parameterization} & \multicolumn{1}{m{33pt}|}{\centering Field of view} & \multicolumn{1}{m{33pt}|}{\centering Camera tilt} & \multicolumn{1}{m{33pt}|}{\centering Camera roll} \\
\hline
Horizon and field of view & 6.061\degree & 2.663\degree & 1.238\degree  \\
Horizon and vertical vanishing point & \textbf{4.911\degree} & \textbf{2.091\degree} & \textbf{0.981\degree} \\
\hline
\end{tabular}
\end{center}
\caption{Comparison of the error in estimated internal and external camera parameters on the CARLA-VP dataset using different parameterization techniques (lower is better). It can be seen that the CNN trained to output the horizon line and the vertical vanishing point gives better performance.}
\label{table:comp_parameterization}
\end{table}

\begin{table}
\small
\begin{center}
\begin{tabular}{|L{85pt}|R{33pt}|R{33pt}|R{33pt}|}
\hline
\multicolumn{1}{|m{85pt}|}{\centering CNN Architectures} & \multicolumn{1}{m{33pt}|}{\centering Field of view} & \multicolumn{1}{m{33pt}|}{\centering Camera tilt} & \multicolumn{1}{m{33pt}|}{\centering Camera roll} \\
\hline
 VGG-M & 6.163\degree & 2.332\degree & 1.534\degree \\
 VGG-16 & 5.385\degree & 1.887\degree & 1.207\degree \\
 Resnet-50 & 4.509\degree & 1.755\degree & 1.104\degree \\
 Resnet-101 & 4.534\degree & 1.652\degree & 1.234\degree \\
 Inception-V1 & 6.773\degree & 2.374\degree & 1.456\degree \\
 Inception-V4 & \textbf{4.130\degree} & \textbf{1.509\degree} & \textbf{0.853\degree} \\
 \hline
\end{tabular}
\end{center}
\caption{Comparison of the error in estimated internal and external camera parameters on the CARLA-VP dataset using different CNN architectures as a component of our pipeline (lower is better).}
\label{table:comp_cnnarchs}
\end{table}

We evaluate this approach to see how it performs against our original
method. The results are shown in table~\ref{table:comp_parameterization}.
We observe that the four scalar parameterization does better in
estimating all the internal and external parameters of the camera.
We believe that one of the major reasons is that the vertical vanishing
point is easier to estimate given that the orientation of the ground
plane or the direction of vertical lines on the ground plane is directly
observable from the image. On the other hand, the camera's field of view
can be difficult to estimate given the fact that two images which are
captured from cameras with different focal lengths and different distances
to the objects may appear very similar.

There are other advantages of our method as well. It is easier to
verify the vertical vanishing point manually from an image. It also
gives us an additional method for calculating the tilt of the camera
and we can average it with the tilt value calculated from the horizon
line. Furthermore, the focal length of the camera is relatively more
sensitive to small errors at  large values of the field of view due to the
$\tan$  relation in~(\ref{eq:focal_length_fov}) (the  focal length is inversely proportional to
$\tan$ of the field of view. Therefore, for large values of the field of view,
a small change in the field of view (e.g.\ change from 115 to 117 compared
to 45 to 47) will cause f to change more since the slope of the tangent
increases very quickly as it  approaches $\pi/2$).

\vspace{-10pt}
\paragraph{Choice of trunk architecture.} We compare the performance using
a number of different and popular CNN architectures. In each case, the CNN is initialized
by pre-training on ImageNet
classification.  We
start with a simple model \ie
\mbox{VGG-M}~\citep{chatfield2014return} with relatively few
parameters,  and then train progressively more complex and deeper
CNNs. Table~\ref{table:comp_cnnarchs} shows the comparison of
the tested networks on the CARLA-VP dataset. We use the best performing
Inception-v4~\citep{szegedy2017inception} architecture for the remaining 
results.

\subsection{Comparison with other methods}
\label{subsec:comparison_third_party}
We compare our method for estimating the horizon line on two public image dataset benchmarks.

\begin{figure}
\centering
\includegraphics[width=0.8\columnwidth]{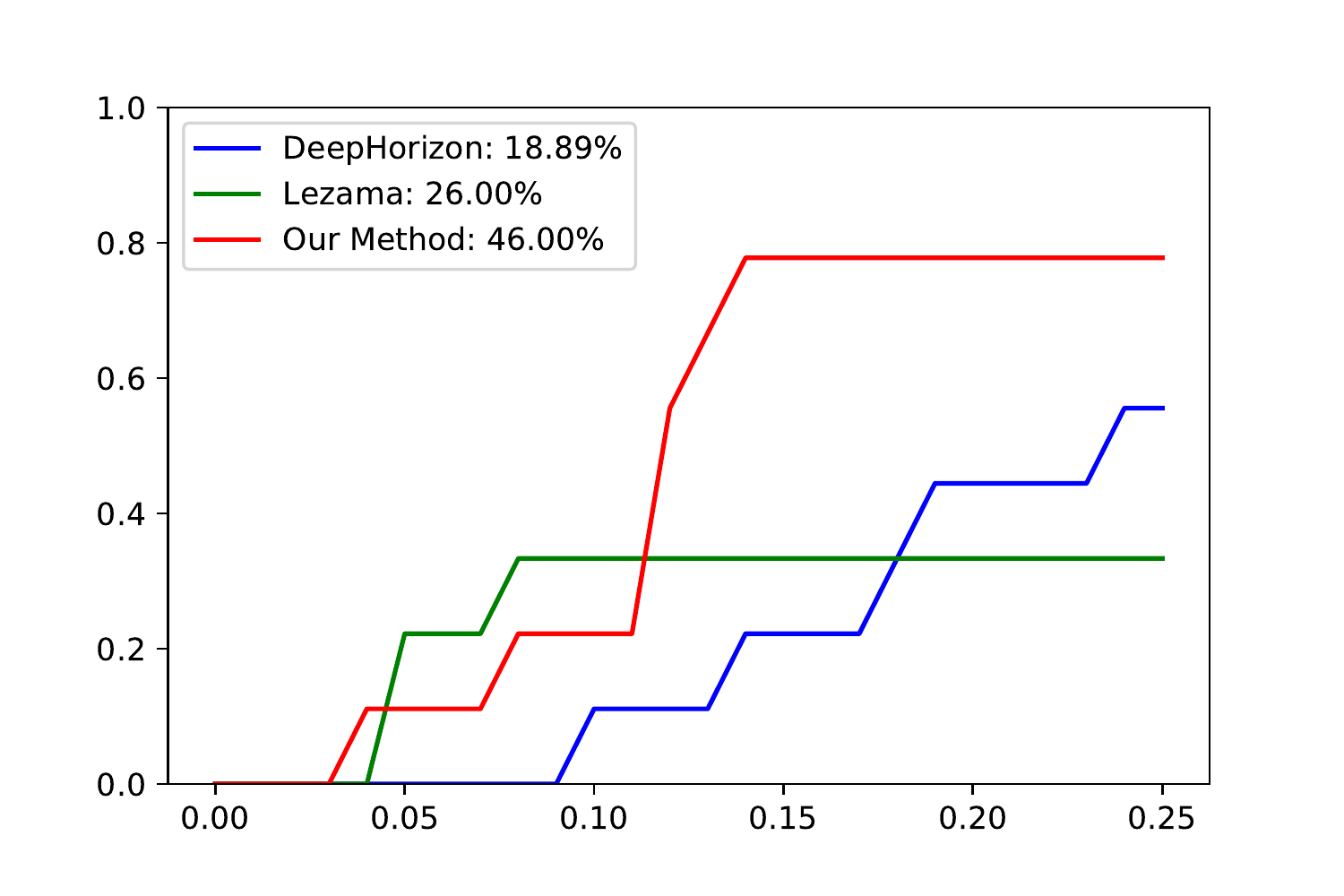}
\caption{\small Horizon line detection AUC performance  on the VIRAT Video dataset. 
Comparison of  our method (trained on synthetic data) against
 DeepHorizon~\citep{workman2016horizon} and Lezama~\citep{lezama2014finding}.
The dataset contains a variety of images with various positions and orientations of the horizon line. Our method does significantly better than the state-of-the-art.}
\label{fig:quant_comp_with_dh}
\end{figure}

\vspace{-5pt}
\subsubsection{Comparison on the VIRAT Video dataset}

\paragraph{The VIRAT video dataset~\citep{virat2011dataset}.} 
This dataset contains videos with fixed cameras
(table~\ref{table:comp_datasets}) along with the corresponding
homography matrices for the ground planes. It also contains object and
event annotations. We use single images extracted from videos in this dataset and
extract the ground truth horizon lines from the given homography matrices
using~(\ref{eq:gt_vanishing_point}).

We compare our method, trained on the synthetic CARLA-VP dataset, to two other
methods:  DeepHorizon~\citep{workman2016horizon} 
using the provided API;  and Lezama~\citep{lezama2014finding}  using the code published by the
authors. Therefore, this dataset is unseen for all three methods.
See figure~\ref{fig:quant_comp_with_dh} for the results. 

We observe that our method outperforms DeepHorizon (state-of-the-art)
and Lezama by a significant margin. Upon closer inspection, we see
that the DeepHorizon method struggles on images where the horizon line
lies outside the image, while our method is able to do well on such
images. One of the reasons could be that DeepHorizon gives good weightage to segmentation between the ground plane and the sky to aid the horizon prediction, but this cue may not be available when the camera is titled significantly.

\begin{table*}
\small
\centering
\begin{tabular}{|L{100pt}|L{131pt}|C{65pt}|R{40pt}|}
\hline
\multicolumn{1}{|m{100pt}|}{\centering Method} & \multicolumn{1}{m{131pt}|}{\centering Datasets} & \multicolumn{1}{m{65pt}|}{\centering Post-Processing} & \multicolumn{1}{m{40pt}|}{\centering AUC} \\
\hline
 \citet{lezama2014finding} & \emph{(requires no training)} & \cmark & 52.59\% \\ 
 \citet{zhai2016detecting} & 110K Google Street & \cmark & 58.24\% \\
 \citet{workman2016horizon} & HLW+500K Google Street & \xmark & 69.97\% \\
 \citet{workman2016horizon} & HLW+500K Google Street & \cmark & 71.16\% \\
 Ours & HLW & \xmark & \textbf{74.52\%} \\
 \hline
\end{tabular}
\caption{\small Horizon line detection AUC performance on the HLW test dataset.
Comparison of our method against other horizon-line detection methods. The datasets column shows the datasets the methods were trained on.}
\label{table:comp_performance_methods}
\end{table*}

\begin{figure}
\centering
\includegraphics[width=0.8\columnwidth]{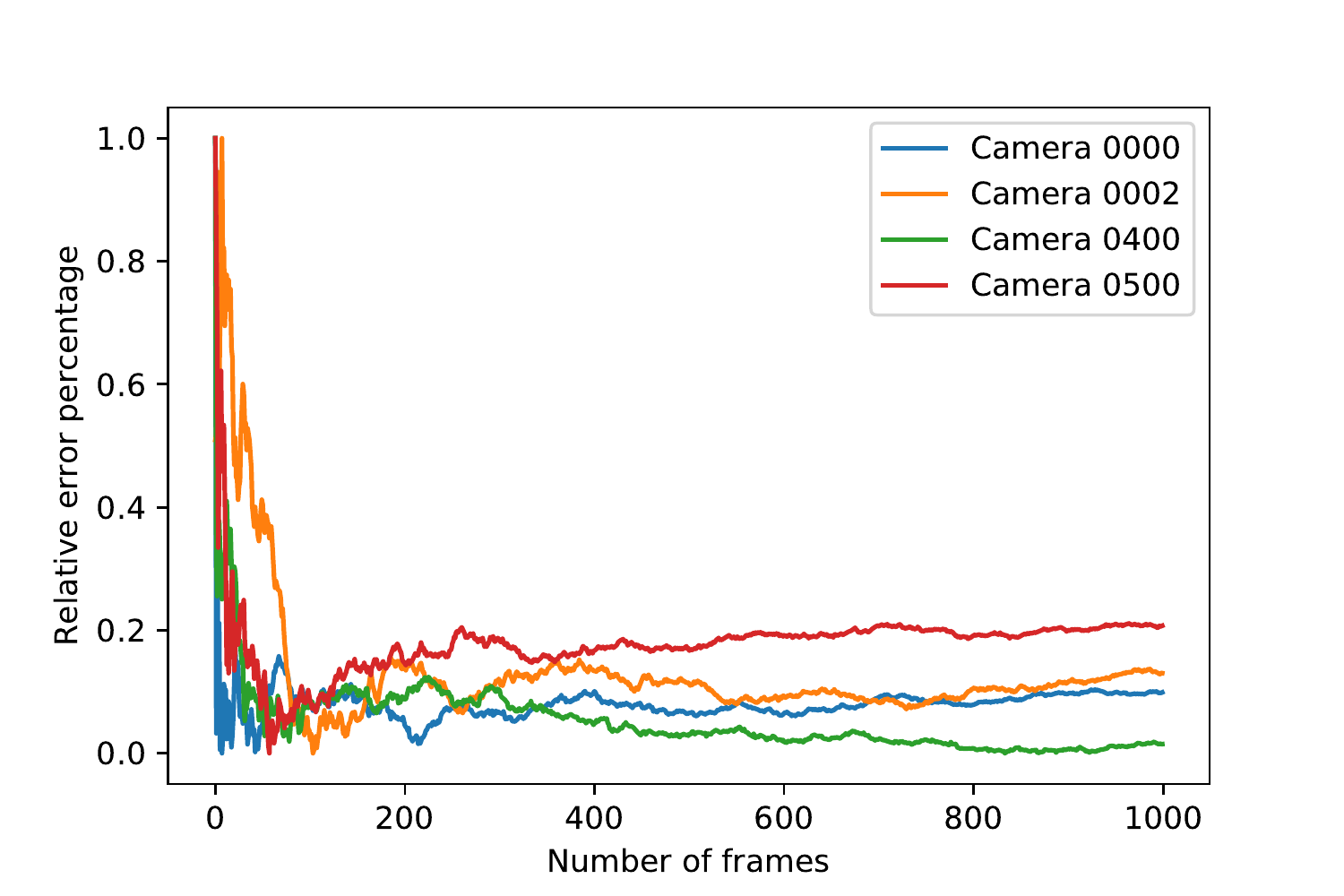}
\caption{Reduction of relative error in estimating the focal length of the camera as the estimates from
more frames are averaged for different videos in the VIRAT Video dataset. We observe that the estimate 
asymptotes at  around 400 frames.}
\label{fig:comp_videos_frames}
\end{figure}

\begin{figure}
\centering
\begin{subfigure}{.5\columnwidth}
  \centering
  \includegraphics[width=0.95\columnwidth]{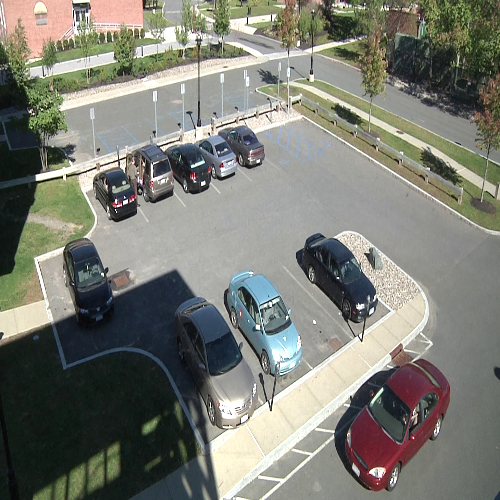}
  \label{fig:sub5}
\end{subfigure}%
\begin{subfigure}{.5\columnwidth}
  \centering
  \includegraphics[width=0.95\columnwidth]{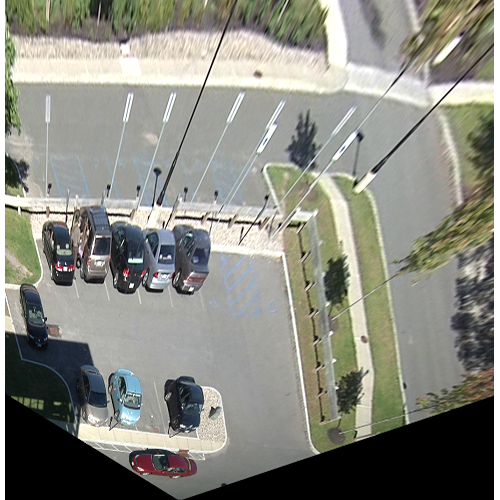}
  \label{fig:sub6}
\end{subfigure}
\begin{subfigure}{.5\columnwidth}
  \centering
  \includegraphics[width=0.95\columnwidth]{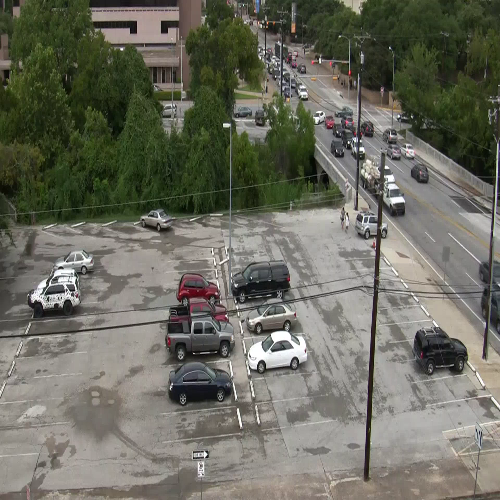}
  \label{fig:sub7}
\end{subfigure}%
\begin{subfigure}{.5\columnwidth}
  \centering
  \includegraphics[width=0.95\columnwidth]{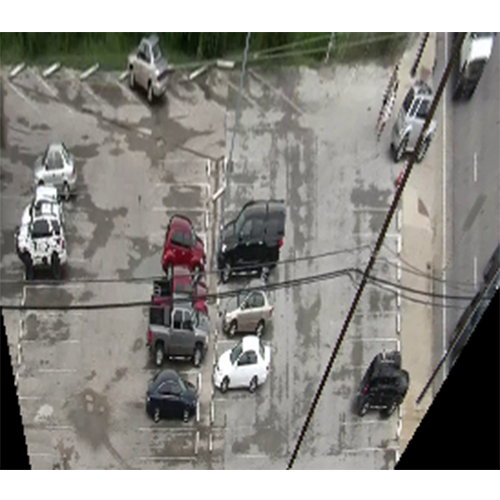}
  \label{fig:sub8}
\end{subfigure}
\begin{subfigure}{.5\columnwidth}
  \centering
  \includegraphics[width=0.95\columnwidth]{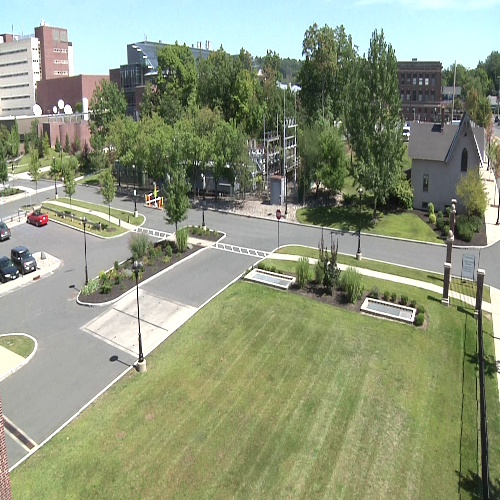}
  \label{fig:sub9}
\end{subfigure}%
\begin{subfigure}{.5\columnwidth}
  \centering
  \includegraphics[width=0.95\columnwidth]{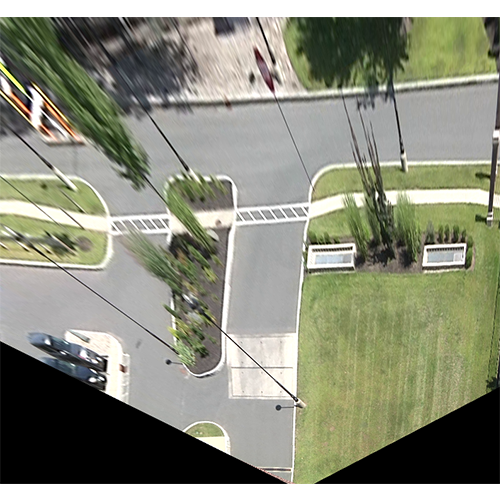}
  \label{fig:sub10}
\end{subfigure}
\caption{(Left) Source images (Right) Bird's eye view of the corresponding source images which are automatically calculated using our method.}
\label{fig:visual_virat}
\end{figure}

We show qualitative results for some of the scenes from the VIRAT
Video dataset in figure~\ref{fig:visual_virat}, which contains the
original images and their corresponding bird's eye views. The obtained
bird's eye views have the correct geometric proportions for different
objects present in the scene such as dimensions of lane markings and roads. This means  that we can obtain
Euclidean measurements in the scene if we know one reference distance in the
image.
We observe that our method is able to transfer well to the real-world
images and generates veridical views.

\vspace{-10pt}
\paragraph{Real time performance on Videos.}
Since our method does not involve any other refinement steps
like expectation maximization \etc as used in~\citep{zhai2016detecting}, it is very fast and takes around
40 ms per image on a lower-middle end GPU (GTX 1050
Ti). This amounts to 25 frames per second, thus making it suitable for
application to videos in real time.

Here, we evaluate a simple approach which can be used to improve the
performance. We apply our method to different videos
from the VIRAT Video dataset and average the values for the
internal and external parameters of the camera (rather than  the
homography matrix). This allows us to refine our estimated values 
continuously and get more reliable and stable results. We observe
that the estimate of the camera parameters gets
more accurate as more frames are averaged from the video. See
figure~\ref{fig:comp_videos_frames} for a visualization of the focal length error. The 
estimated value for the focal length approaches
the ground truth value as the number of
frames increases.

\vspace{-10pt}
\subsubsection{Comparison on the HLW Dataset}

In this section, we present our results on the latest horizon
detection dataset known as \emph{Horizon Lines in the Wild} (HLW). 

\vspace{-10pt}
\paragraph{The Horizon Lines in the Wild (HLW) dataset~\citep{workman2016horizon}.} 
This dataset contains around 100K
images with ground truths for the horizon line. The dataset
mostly contains images with a very small tilt or roll of the camera
and the camera is close to a ground person height. This causes the
horizon line to be visible in most of the images.

We use pre-initialized weights from ImageNet to train our method on
the training set of the HLW dataset to compare
with other methods. See table~\ref{table:comp_performance_methods} for a
summary of performance of different methods on the HLW test set. We
achieve 74.52\% AUC, outperforming the previous state-of-the-art
method~\citet{workman2016horizon} with a relative improvement of 4.72\%.

Our network predicts the geometry in one forward pass, without any
kind of post-processing involved. Compared to
this,~\citet{lezama2014finding} detect line segments in the image
initially, and compute vanishing points from them which gives the
horizon line.~\citet{zhai2016detecting} estimates horizon line
candidates from the CNN. Then they estimate the zenith vanishing point
using these horizon lines. Based on this, they estimate the horizontal
vanishing points on the horizon line candidates and select the horizon
line with maximum score.~\citet{workman2016horizon} estimate the
horizon line directly from the image using a CNN, but they use further
post-processing techniques to achieve their best results.

\vspace{-5pt}
\section{Conclusion}
We have presented a complete pipeline for removing perspective
distortion from an image, and obtaining the bird's eye view from a
monocular image automatically under geometric constraints. Our method can be used as plug and play
to help other networks which suffer from multiple-scales due to
perspective distortion such as vehicle tracking~\citep{o2011vision},
crowd counting~\citep{liu2018geometric,liu2018context} or penguin
counting~\citep{arteta2016counting} \etc. Our method
is fast, robust and can be used in real-time on videos to generate a
bird's eye view of the scene.

Note, the finite points used to represent the geometric entities in this work do not correspond directly to observable features in the image. A possible improvement in future work would be to design a projection method so that they do correspond.

\vspace{3pt}
\noindent\textbf{Acknowledgements:}  
\small We would like to thank Nathan Jacobs for his help in sharing
the DeepFocal~\citep{workman2015deepfocal} dataset.
We are grateful for financial support from the EPSRC Programme Grant Seebibyte
EP/M013774/1. 

{\small
\bibliographystyle{IEEEtranSN}
\bibliography{refs}
}

\end{document}